\pdfoutput=1
\PassOptionsToPackage{table,dvipscolor}{xcolor}
\documentclass[11pt]{article}

\usepackage[final]{acl}

\usepackage{times}
\usepackage{latexsym}

\usepackage[T1]{fontenc}

\usepackage[utf8]{inputenc}

\usepackage{microtype}

\usepackage{inconsolata}

\usepackage{graphicx}

\title{Simultaneous Translation with Offline Speech and LLM Models in CUNI Submission to IWSLT 2025}

\author{Dominik Macháček \and Peter Pol{\' a}k 
  \\ \\
Charles University, Faculty of Mathematics and Physics, \\ 
Institute of Formal and Applied Linguistics, Czech Republic \\
\texttt{\{machacek,polak\}}\texttt{@ufal.mff.cuni.cz} 
}

\usepackage{xspace,mfirstuc,tabulary}

\usepackage{inconsolata}

\usepackage{graphicx}
\usepackage{xurl}
\usepackage{csquotes}
\usepackage{twemojis}
\usepackage{tabularray}
\usepackage{tikz}
\usepackage{amsmath}
\usepackage{pgfplots}
\pgfplotsset{compat=1.18}
\usepackage{subfigure}
\usepackage{caption}
\usepackage{soul}
\usepackage{multirow}

\usepackage[capitalize,noabbrev]{cleveref}

\def\XXX#1{} 

\definecolor{lightblue}{RGB}{66,133,244}

\date{}

\begin{document}
\maketitle
\begin{abstract}

This paper describes Charles University submission to the Simultaneous Speech Translation Task of the IWSLT 2025. We cover all four language pairs with a direct or cascade approach. The backbone of our systems is the offline Whisper speech model, which we use for both translation and transcription in simultaneous mode with the state-of-the-art simultaneous policy AlignAtt. We further improve the performance by prompting to inject in-domain terminology,  and we accommodate context. Our cascaded systems further use EuroLLM for unbounded simultaneous translation. Compared to the Organizers' baseline, our systems improve by 2 BLEU points on Czech to English and 13-22 BLEU points on English to German, Chinese and Japanese on the development sets. Additionally, we also propose a new enhanced measure of speech recognition latency.

\end{abstract}

\def\SimulStreaming{SimulStreaming}

\section{Introduction}


In this paper, we describe the submission of the Charles University (CUNI) system to IWSLT 2025 Simultaneous Speech Translation Task \cite{abdulmumin-etal-2025-findings}. 
Our system is built on top of Whisper \cite{Whisper-paper} with AlignAtt \cite{papi23_interspeech} simultaneous policy. To achieve higher translation quality, we apply beam search and prompting for in-domain terminology. In our end-to-end system for the Czech-to-English translation, we also exploit previous translations as a context. For the translation into German, Chinese, and Japanese, we adopted a cascaded approach consisting of Whisper for English ASR and EuroLLM \cite{eurollm} for translation. We validate our systems' latency in computationally unaware simulation. Our Czech-to-English systems work both in 2-second and 4-second latency regimes required by IWSLT 2025 (``low'' and ``high''). The English-to-German, Chinese and Japanese systems are available only in the high-latency regime of 4-5 seconds. For an overview of our systems, see \Cref{tab:overview}.


Our main goal in this submission is to create a robust and straightforward implementation that can be used in further research as well as in many realistic use cases. We name the implementation \textbf{\SimulStreaming{}} and publish it at 
\centerline{\url{https://github.com/ufal/SimulStreaming}.}

Among the strengths of our submitted system is very high quality, because of using high-performing foundation models, and very high multilinguality. Whisper allows direct translation from 99 speech source languages to English, and EuroLLM allows English translation into 35 languages. Our systems are also adaptable; the prompts and in-context learning allow injecting specific in-domain terminology.

Moreover, although we primarily focus on computationally unaware latency, our system is practically usable in real time only with feasible hardware resources. It requires hosting Whisper large 1.6B parameters model and  and the EuroLLM 9B parameters model. 

Our second goal is to evaluate the state-of-the-art methods in combination. Our results show improvements by 2 BLEU points on Czech to English over the organizers baseline, and 13-22 BLEU points on English to German, Chinese, and Japanese, which highlights the effectiveness of our systems.

We conclude we have reached both goals. The original contributions of this work is the \SimulStreaming{} implementation and evaluation, and also a new enhanced method for measuring ASR latency using Continuous Levenshtein Alignment (see \Cref{sec:latency}).

\def\yes{yes}
\def\no{no}
\def\na{-}
\begin{table*}[]
    \centering
    \begin{tabular}{cl|c|c}
       &  & \textbf{Cs-En} & \textbf{En-\{De,Zh,Ja\}} \\
       \hline
\multirow{6}{*}{\textbf{Speech-to-text}} &  model & Whisper large-v3 &  Whisper large-v3       \\
& task & translate to En & transcribe En \\
& beam & \yes{}     & \no{} \\
& prompt &  \yes{}  & \no{} \\
& context &  \yes{} & \no{} \\
& simult. policy & AlignAtt & AlignAtt \\
\hline
\multirow{4}{*}{\textbf{+ Text-to-text}} 
&  model & \na{} &  EuroLLM-9B-Instruct \\
& prompt & \na{} & \yes{} \\
& context & \na{} & \yes{} \\
& simult. policy & \na{} & LocalAgreement \\
\hline
\multirow{2}{*}{\textbf{Latency regime}} 
& 2 seconds (``low'') & \yes{} & \no{} \\
& 4-5 seconds (``high'') & \yes{} & \yes{}
    \end{tabular}
    \caption{Overview of CUNI systems submissions to IWSLT 2025 Simultaneous Speech Translation Task.}
    \label{tab:overview}
\end{table*}




\section{Background}

\def\highlighted#1{\textbf{#1}}


\highlighted{Whisper} is among the top-performing ASR and speech translation models for 99 languages. It has the ability to use initial and context prompts, which makes it adaptable for in-domain terminology. Works such as \citet{machacek-etal-2023-turning} and \citet{simul-whisper} show that Whisper is adaptable to simultaneous mode, although it is primarily designed for offline mode. 
Whisper is available in multiple model versions that differ in size and quality. We use the large-v3 model, which achieves the highest quality.

\highlighted{AlignAtt} \cite{papi23_interspeech} is a simultaneous policy. Given an offline translation model, partial source and previous target, it detects where to stop generating the partial target, which is when the most attended source frame by the decoder is behind a threshold. \citet{papi23_interspeech} shows that this policy outperforms all previously proposed policies. \citet{simul-whisper} later showed that AlignAtt also works with Whisper. 

\highlighted{LocalAgreement} \cite{polak-etal-2022-cuni,polak23_interspeech} is a simultaneous policy that considers the target prefixes of two subsequent updates, each processing a newly incoming source chunk. It emits their longest common prefix as confirmed and uses it in forced decoding of the latter chunks.

\highlighted{Simul-Whisper} \cite{simul-whisper} is an implementation of the simultaneous mode with Whisper using AlignAtt. It is an extension of the original OpenAI Whisper inference using the Torch deep learning framework. Simul-Whisper supports only ASR of speech that is segmented into individual sentences, and computationally unaware simulation, while the IWSLT 2025 Simultaneous Task focuses on a more realistic case of unbounded speech \cite{papi-etal-2024-how-real} without any explicit sentence boundaries. On the other hand, \highlighted{Whisper-Streaming} \cite{machacek-etal-2023-turning} is our implementation of Whisper with the LocalAgreement simultaneous policy and reprocessing the audio buffer from the beginning with every incoming source chunk, which is less computationally effective than AlignAtt. On the other hand, Whisper-Streaming supports both computationally aware and unaware simulations, as well as unbounded speech. It integrates Silero VAD \cite{Silero_VAD} that incrementally detects silence and non-voice sounds vs.\ voice.


\highlighted{EuroLLM} \cite{eurollm} is a recent large language model for text-to-text translation between 35 EU and non-EU languages, including English, German, Japanese, and Chinese. It is a decoder-only model of the LLaMA family. We use its 9B parameter version with instruction tuning. It supports a system prompt, which can be used to suggest the domain, and a maximum context of 4096 tokens, which spans over 10 minutes of English source and German target of ACL 6060 \cite{acl6060} dev set reference (see estimation in \Cref{sec:maxcontext}). We use EuroLLM with the fast inference framework CTranslate2. It enables fast computation and efficient memory usage; however, it currently does not provide access to attention weights. Therefore, we can not apply the AlignAtt policy to EuroLLM, so we use it with the LocalAgreement simultaneous policy, 
which is the best-performing policy that does not require attention weights.




\section{Direct Simultaneous Speech-to-Text with Whisper and AlignAtt}

Let us describe the process of simultaneous speech-to-text processing, which we apply to direct Czech-to-English translation and to English ASR in the cascaded system for English to German, Chinese, and Japanese.

Our system uses our implementation called \SimulStreaming{} which merges the Simul-Whisper AlignAtt policy with the Whisper-Streaming interface to support unbounded speech processing. Moreover, we extend the original Simul-Whisper with the following enhancements. First, we added support for the Whisper large-v3 model. Second, in addition to transcription, we enabled translation. Then, to improve quality, we implement beam search decoding. Finally, we incorporate support for initial prompts and contextual information from the preceding audio buffers. 

\def\MinChunkSize{\textit{MinChunkSize}}
\def\BufferLen{\textit{BufferLength}}
\def\MaxContextLen{\textit{MaxContextLength}}
\def\StaticPrompt{\textit{StaticPrompt}}
\def\Frames{\textit{Frames}}
\def\Beams{\textit{Beams}}
\def\NonStaticPrompt{\textit{NonStaticPrompt}}
\def\Prompt{\textit{Prompt}}

Our simultaneous speech-to-text pipeline consists of the prototypical processing steps that are described in \citet{papi-etal-2024-how-real}, Section~3.1.

\begin{enumerate}
    \item Audio acquisition. 
    \item Audio segmentation. Silero Voice Activity Detection (VAD) iterator with the same default parameters as in Whisper-Streaming is applied (minimum chunk size 0.04 seconds, minimum non-voice duration 500 ms, voice is padded with 100 ms). When VAD detects non-voice in the 0.04-second chunk, the chunk is discarded. When VAD detects voice, it holds it until \MinChunkSize{}\footnote{We mark the system parameters that we tune with italics.} seconds of voiced audio accumulate, or until the end of voice is detected. The accumulated voiced audio is passed to the next step.
    
    \item Speech buffer update. The incoming chunk, which has \MinChunkSize{} seconds if the end of voice is not detected, or less otherwise, is concatenated with the speech buffer.
    \item Hypothesis generation. Whisper large-v3 model encodes the speech buffer and populates the decoder's Key-Value cache with the representation of the optional initial prompt, previous context, and forced-decoded target prefix. Then the model decodes the target as long as the AlignAtt policy allows. 
    If the current chunk is not final, the decoding continues until the most attended source frame is close to the end of the audio, which is indicated by the \Frames{} parameter. In our proposed beam search implementation, we decode until the top beam hypothesis is attended behind the threshold. In case the current chunk is final, we decode until the last 4 frames, as the Simul-Whisper authors propose. 
    
    
    \item Buffers selection. There are the following four buffers in our implementation: (1) source audio buffer, (2) forced decoding target buffer that contains the stable part of the hypothesis that was decoded from current audio buffer, (3) context buffer, which is the transcript or translation from the audio segments that were pushed away from the audio buffer, and (4) initial prompt, for example a text that can contain terminology or initiate the style of decoding.
    
    If the audio buffer has the length of \BufferLen{} seconds or more, we remove the first speech chunk from the source audio buffer. At the same time, we move the text that was decoded with the first chunk from the forced decoding to the context buffer. If the initial prompt and context are longer than \MaxContextLen{}, we trim the complete words from the beginning. A parameter \StaticPrompt{} specifies whether the initial prompt is pushed away with the context or not. 
    
    If finalization is triggered, that is, when the source recording is finished, or when the end of voice was detected, the buffers are cleared. 

\end{enumerate}

\XXX{tady by se mohla hodit tabulka, která by vysvětlila všechny parametry. Ale nestíháme, nebude. \MinChunkSize, \BufferLen, \MaxContextLen, \StaticPrompt, \Frames, \Beams}




\section{Simultaneous Translation with EuroLLM}

                
We implement EuroLLM simultaneous translation using a chat template. We design a system prompt asking the model to perform simultaneous interpreting at a conference and specifying the translation direction. The chat is followed by the user's message containing the source prefix, and by the assistant's reply that contains the previous target prefix to continue. The chat is initiated with one sentence pair as an in-context example because we observed that without that, the model tends to produce text that is not a translation, especially for a short source.


Our simultaneous translation consists of steps that are analogous to the prototypical ones in \citet{papi-etal-2024-how-real}:

\def\BufferTrimmingStrategy{\textit{BufferTrimmingStrategy}}
\def\BufferTrimmingStrategies{\textit{BufferTrimmingStrategies}}
\def\Sentences{\textit{Sentences}}
\def\Segments{\textit{Segments}}

\begin{enumerate}
    \item Source acquisition: The punctuated text produced by the simultaneous Whisper English ASR.
    \item Segmentation: Because we assume computational unaware mode and English as the source language, we segment the source into individual words by spaces. A parameter \MinChunkSize{} specifies the number of new words in each update.
    \item Buffer update: The newly incoming source words are appended to the previous source.
    \item Buffer trimming: In our initial experiments, we observed that the model tends to hallucinate with larger context. Therefore, we trim the source-target buffer if it has more tokens than \MaxContextLen{}. We apply one of the two buffer trimming strategies:
    \begin{enumerate}
        \item \Sentences: Detect sentences by punctuation in the source and target, trim the first sentence in the source and target, while the context length is too large and there is at least one sentence left in each buffer. This strategy assumes that there is a one-to-one correspondence between the source and target sentences. This strategy seems to be sufficient for English-to-German translation, but not for English to Chinese and Japanese.
        \item \Segments: The source-target buffer contains the source-target pairs as they were received and generated, including empty targets. If the buffer is too long, one pair is trimmed. Although the buffer is not completely parallel and the source is typically more ahead, this strategy appears to be more optimal for English-to-Chinese and Japanese translations than \Sentences{}. Additionally, this strategy does not require processing a slow parallel word-alignment model.
    \end{enumerate}
    \item Hypothesis generation: The source and target buffers are transformed into the chat tokens as described above. EuroLLM's reply is generated. The new target prefix is compared to the one from the previous update, and their longest common prefix (LocalAgreement policy) is considered as a newly confirmed hypothesis. The unconfirmed hypothesis suffix is held for confirmation with the following update.
\end{enumerate}

\section{Development}

\paragraph{Dev sets} For English-to-German, Chinese, and Japanese translations, we use the ACL6060 development set as provided by the IWSLT organizers. For Czech-to-English translation, we use the IWSLT 2025 dev set. However, we found that the ParCzech subset is segmented in this dev set, while the 2025 test set will be unsegmented. Therefore, we merged the subsequent segments from the same speech. Since there were two very diverse subsets, ParCzech and Robothon, we selected the final candidate based on the average of quality scores on the merged ParCzech and Robothon. Finally, to meet the shared task conditions, we filtered out the candidates that did not meet the latency criteria on the unsegmented dev set.

\paragraph{MT metric} We selected the primary candidates using ChrF because ChrF tends to be more reliable than BLEU in simultaneous translation \cite{machacek-etal-2023-mt}.

\paragraph{Translation Latency} For translation latency, we use the StreamLAAL metric (SLAAL, \citealp{papi-et-al-2024-streamatt}) as proposed by IWSLT organizers. For English ASR latency, we use the following algorithm.

\subsection{ASR Latency with Continuous Levenshtein Alignment}
\label{sec:latency}

We propose an improvement of the algorithm for the average word latency of the ASR. We call it ``ASR Latency with Continuous Levenshtein Alignment.'' The improvement over existing methods stems from (1) the more accurate character-level alignment and (2) minimum edit distance alignment that prefers continuous sequences of edit operations to prevent coincident alignment to deleted or inserted segments that would contribute to unrealistic latency. The algorithm is as follows.

Assume a gold transcript with word-level timestamps and an ASR transcript where each word is assigned its emission time.

First, create a dynamic programming matrix for the Levenshtein minimum edit distance alignment of the gold and ASR transcript at the character level. Character-level alignment is more accurate than word-level because it is more robust to minor deviations from the gold transcripts, such as suffixes, when the other part of the word is correct. The disadvantage is computation and memory complexity, which is quadratic, and therefore much higher on characters than on words. However, we were able to compute roughly 12 minutes of transcripts in a feasible time. Longer transcripts can be segmented.

Second, when generating the minimum distance alignment, prioritize continuous sequences of Copy or Substitute operations over interruptions with Deletes or Inserts. The illustration is in \Cref{fig:alignments}. The reason is to prevent alignment to deleted segments too far ahead or behind, which would lead to incorrect latency.

Third, convert aligned characters to the sequence of aligned words. Fourth, for each word in the transcript that is aligned to any gold word, estimate its latency as the word's emission time minus the timestamp of its gold-aligned word. Fifth, report the average word latency.

\begin{figure}
        \centering
        \begin{tabular}{l|c@{~}c@{~}c@{~}c@{~}c@{~}c@{~}c@{~}c@{~}c@{~}c}
         gold    & t & h & e & \textvisiblespace & t & a & b & l & e \\
    \hline
    interrupted alignment  & t &   &   &   &   & a & b & l & e \\
   edit operations        & C & D & D & D & D & C & C & C & C \\
   \hline
   continuous alignment &   &  &  &  & t & a & b & l & e \\
edit operations        & D & D & D & D & C & C & C & C & C \\
        \end{tabular}
    \caption{Illustration of interrupted vs.\ continuous alignment of the ASR candidate ``table'' to the gold ``the table''. Both alignments have identical edit distance (4 Deletions, 5 Copies), but the bottom one includes a longer continuous sequence of Copies. The interrupted alignment is incorrect.}
    \label{fig:alignments}
\end{figure}

We publish an implementation of the ASR Latency with Continuous
Levenshtein Alignment at

\centerline{\url{https://github.com/ufal/asr_latency}.}

\section{Results}

\subsection{Czech-to-English Translation}

For the Czech-to-English translation, we use the direct speech translation with Whisper and AlignAtt policy.
First, we investigate the impact of \Beams{} and \BufferLen{}. For that, we use a 30-minute subset of the merged ParCzech dev set. 

\paragraph{Beam search} We set the \MaxContextLen{} to 0, \BufferLen{} to 25 seconds, \Frames{} threshold to 4, and \MinChunkSize{} to 3 seconds.
\Cref{tab:beam} contains MT quality scores with different \Beams{}. We observe a ChrF score improvement by 1.04 with 5 beams compared to 1. With \Beams{} 4 and 8, we observe analogous gains, with maximum at 5 beams. The latency (SLAAL) decreases negligibly with higher beams. 

\begin{table}[]
    \centering
    \begin{tabular}{l|rrrrr}
Beams & 1 & 2 & 6 & 5 \\
\hline
ChrF & 48.03 & 48.77 & 48.89 & \textbf{49.07} \\
SLAAL & 2373 & 2393 & 2308 & 2285 \\
    \end{tabular}
    \caption{Impact of beam search on MT quality (ChrF) and latency (SLAAL) in milliseconds.}
    \label{tab:beam}
\end{table}

\paragraph{Buffer Length} Then, we investigated the \BufferLen{} parameter. The setup is the same as with beam search, except that we set \MinChunkSize{} to 1.75 seconds and \Frames{} to 4. \Cref{tab:buffer} shows the results. Maximum quality is with \BufferLen{} 30 seconds. We observe analogous results with \Frames{} set to 80.

\begin{table}[]
    \centering
    \begin{tabular}{l|r@{~}r@{~}r@{~}r@{~}r@{~}r@{~}r@{~}}
\BufferLen{}& 15 & 20 & 25 & 28 & 30 \\
\hline
ChrF & 47.65 & 48.09 & 48.24 & 48.67 & \textbf{48.78} &   \\
SLAAL & 2698 & 2627 & 2830 & 2920 &  2928 \\
    \end{tabular}
    \caption{Impact of \BufferLen{} on MT quality (ChrF) and latency (SLAAL) in milliseconds.}
    \label{tab:buffer}
\end{table}


Therefore, we further set 
\BufferLen{} to 30 seconds.

\paragraph{Grid search} Then, we perform grid search to find the optimal \MinChunkSize{}, \Frames{}, and \Beams{} parameters to meet the low-latency threshold of the IWSLT 2025 Simultaneous task, which is below SLAAL 2000ms, and the high-latency threshold below 4000ms SLAAL. For that, we used the merged ParCzech and Robothon portions of the dev set, and we averaged their ChrF score. We found 4 candidates for the low-latency regime that were near 2000 SLAAL. Their scores are in \Cref{tab:grid}. For high latency, we selected one candidate.

\begin{table}[]
    \centering
    \begin{tabular}{ll|ll}
\MinChunkSize{} & \Frames{} & ChrF & SLAAL \\
\hline
  1.2 & 25 & 49.72 & 1715 \\
  1.4 & 35 & 49.75 & 2091 \\
  1.6 & 25  & 49.83 & 2166 \\
  1.4 & 30  & 49.83 & 2067 \\
  \hline
  1.8 & 25 & 49.93 & 2636 \\
    \end{tabular}
    \caption{Pre-selected top candidates for low latency (upper part of table, with SLAAL near 2000) and high latency (lower part) by average ChrF on the merged ParCzech and Robothon portions of the dev set. All of them are with 2 \Beams{}.}
    \label{tab:grid}
\end{table}

\paragraph{Prompt and Context}
We experimented with \MaxContextLen{}, which can be between 0 and 255 tokens, as Whisper's documentation suggests, and \Prompt{}, which can be any text that initiates decoding. Moreover, the prompt can be \StaticPrompt{}, which stays at the beginning of decoding for all buffers, or \NonStaticPrompt{}, which means it is pushed away by context that reaches maximum length. 

We optimize for two subsets of the Czech-to-English test set in IWSLT 2025: the native ParCzech subset, and the non-native subset for which we have no other information. We assume that the ParCzech are speeches from the plenary sessions of the Chamber of Deputies, Parliament of the Czech Republic, similarly to the dev set. We noticed a specific terminology that the Whisper model is not aware of. For example, the terms ``Chamber of Deputies,'' ``deputy,'' and ``chairman'' are often missing. They are alternated with terms such as ``Senate'', ``MP'', ``Ambassador'', and ``President'', which are wrong in the ParCzech domain. Therefore, we attempted to inject these terms via prompting. 

We proposed 13 prompts, 9 of which were specific to the ParCzech, and 4 of them were general, applicable to any domain. We evaluated these prompts on the ParCzech portion of the dev set with all static or non-static prompts and varying context lengths. We discovered that half of the prompts increased the performance over the baseline, while the other half decreased it. 

In the end, the prompt ``This is Chamber of Deputies.'' reached the highest quality score. We use this as the static prompt for the ParCzech domain with \MaxContextLen{} 20 for high latency, where it increased ChrF by 0.45, and  with \MaxContextLen{} 250 for low latency, where it increased ChrF by 0.61. For the general domain, we use no context and no prompt for the high latency, and the non-static prompt ``He starts.'' with \MaxContextLen{} 250 for the low latency, as it gained 0.24 ChrF improvement. \Cref{tab:prompt} summarizes our observations.

\begin{table}[]
    \centering
    \begin{tabular}{l|r|r}
         & \textbf{low} & \textbf{high} \\
         \hline
\textbf{baseline} ChrF  &  49.67 & 49.78 \\
context & 0 & 0 \\
prompt & - & - \\
\hline
ChrF  & \textbf{50.28} & \textbf{50.23} \\
context   &  250 & 20 \\
prompt    & ParCzech, static & ParCzech, st. \\
\hline
ChrF & 49.91 & 49.78 \\
context & 250 & 0 \\
prompt & general, non-st. & -

    \end{tabular}
    \caption{ChrF on the merged ParCzech portion of the dev set with the top performing prompt and context setup with the prompt adapted to the ParCzech domain (middle section of the table), or a general prompt (lower section).}
    \label{tab:prompt}
\end{table}

\paragraph{Comparison to  the IWSLT 2025 Baseline}

Although we do not consider validation on the segmented dev set as the most relevant for evaluation on unbounded speech, which is the primary objective in the IWSLT 2025 Simultaneous task, we validate our primary candidate on the segmented dev set and compare it to the IWSLT 2025 organizers' baseline. Their system that reached the highest BLEU score while having SLAAL below 4000 (high latency) was a cascade system with Whisper ASR and M2M100 MT \cite{m2m100}. Similarly, their top-scoring system for the low-latency regime was the SeamlessM4T \cite{seamless2023} direct speech translation model with VAD Segmenter.
The scores are compared in \Cref{tab:baseline}. 

We conclude that in the Czech-to-English simultaneous translation, we outperformed the organizers' baseline by 3.3 BLEU in the low latency and by 2.2 BLEU in the high latency regime.

\begin{table}[]
    \centering
    \begin{tabular}{l|rrr}
& BLEU & ChrF & SLAAL \\
\hline
 baseline low & 15.16 & - & 1777 \\
our low &  \textbf{18.49} & 48.51 & 1763 \\ 
\hline
baseline high & 16.63 & - & 3996 \\
our high & \textbf{18.83} & 48.86 & 2630 \\
    \end{tabular}
    \caption{Comparison of IWLST 2025 organizers' baseline on the segmented Czech-to-English dev set.}
    \label{tab:baseline}
\end{table}

\subsection{English ASR}

We use Whisper with AlignAtt for simultaneous English ASR. We set \BufferLen{} to 30 seconds (maximum). Since Whisper with AlignAtt reached very high quality on the English ASR of the ACL6060 domain with no prompt and no context, we did not attempt to improve it with prompt or context. We perform a grid search for the parameters \MinChunkSize{}, \Frames{}, and \Beams{}. Unlike for Czech to English, \Beams{} set to 1 performed the best in this case of English ASR. We validate with ACL6060 English dev set in computational unaware mode, measuring latency with the algorithm in \Cref{sec:latency}.

Meanwhile, we validated the simultaneous translation of English to German with the gold transcripts. Given the minimum translation latency, we determined the span of latency for ASR in the cascade to fit the high latency regime of IWSLT 2025 Simultaneous task. We selected the top-performing ASR systems from the grid search with various latency levels, each roughly 100 milliseconds from the others. The scores are in \Cref{tab:asr}. We observe very high ASR quality, around 5\% CER (character error rate.

When we looked at the differences in the ASR and gold transcripts, we noticed that the differences are often not errors but a consequence of unspecified orthographical conventions, for example, swapping numerals and digits, capitalization of titles, use of quotation marks, etc. We also noticed that named entities and acronyms tend to be more often incorrect with small \MinChunkSize{} than with large. This is an expected consequence of shorter context. 

\begin{table}[]
    \centering
    \begin{tabular}{r|r@{~}r|r@{~}r@{~}r}
ref.\ & Chunk & Frame & WER & CER & latency  \\
\hline
\#00 & 0.05 & 4 & 14.22 & 5.10 & 494\\
\#10 & 0.15 & 4 & 13.33 & 4.75 & 596\\
\#11 & 0.25 & 15 & 13.10 & 4.66 & 754\\
\#12 & 0.25 & 20 & 13.09 & 4.63 & 845\\
\#20 & 0.5 & 10 & 13.40 & 4.81 & 1037\\
\#21 & 0.5 & 15 & 13.35 & 4.71 & 1149\\
\#22 & 0.5 & 20 & 13.06 & 4.64 & 1262\\
\#23 & 1.4 & 15 & 12.98 & 4.76 & 1389\\
\#24 & 1.5 & 10 & 12.92 & 4.85 & 1461\\
\#25 & 1.4 & 25 & 12.77 & 4.76 & 1522\\
\#30 & 2.0 & 20 & 12.69 & 4.77 & 2143\\
    \end{tabular}
    \caption{Selected top performing English ASR candidates with various latency levels. We report \% WER (range 0\%-100\%, the lower, the better), \% CER, and latency in milliseconds on ACL6060 English dev set. The first column ``ref.'' is a reference under which we will refer this ASR candidate.}
    \label{tab:asr}
\end{table}

\subsection{English to German, Chinese, and Japanese}

The text-to-text simultaneous translation component of our cascade has the parameters \MinChunkSize{}, \MaxContextLen{}, and \BufferTrimmingStrategy{}. We do not tune the system prompt nor the in-context example because we do not presume to have any further background information about the content to be translated.

\paragraph{Latency regime} First, we processed English-to-German translation with gold ASR. We realized that the lowest possible latency, with \MinChunkSize{} 1 and \MaxContextLen{} 300, is 2471 SLAAL, which means that we can not fit under the low-latency threshold of 2000 SLAAL. We can target only the high-latency regime that requires SLAAL under 4000. Furthermore, we observed a lower quality score with \MaxContextLen{} 500 than with 300, and even lower performance with longer context due to hallucinations, mostly repetitions of long sentences. The results are summarized in \Cref{tab:gold-ende}.

\begin{table}[]
    \centering
    \begin{tabular}{r|rrr}
context & BLEU & ChrF & SLAAL \\
\hline
300 & 39.84 & \textbf{67.88} & 2472 \\
500 & 39.71 & 67.44 & 2461 \\
700 & 16.24 & 48.54 & $<$ 0 \\
1000 & 34.51 & 61.35 & $<$ 0 \\
    \end{tabular}
    \caption{English to German scores with gold ASR input, \MinChunkSize{} 1, and various \MaxContextLen{}. The scores are on ACL6060 dev set with 5 documents. SLAAL scores less than zero ($<$ 0) indicate hallucinations in at least one document.}
    \label{tab:gold-ende}
\end{table}

\paragraph{Buffer Trimming Strategy} We observed many hallucinations with English-to-Chinese and Japanese with the buffer trimming strategy \Sentences{}, because the assumption of matching number of source and target sentences was wrong. The buffer often contained only one source sentence and many short sentences in Chinese or Japanese. However, there were no hallucinations when we applied the \Segments{} strategy instead.

\paragraph{Primary Candidates}
Finally, we performed a grid search with \MinChunkSize{}, \MaxContextLen{}, and the ASR candidates, and found the best ChrF scoring setup on the dev set that met the high-latency criterion. See results in \Cref{tab:mt}, where we also compare to contrastive systems and the organizers' baseline. 

We observe high improvement on each language pair, nearly 13 BLEU points on English to German, 22 BLEU on Chinese, and 18 BLEU point on Japanese. We presume that the baseline was not very strong, likely due to hallucinations of the SeamlessM4T model.

\begin{table*}[]
    \centering
    \begin{tabular}{ll|rrr|r}
   &   & \textbf{BLEU} & \textbf{ChrF} & \textbf{SLAAL} & ASR latency \\
  \hline
\multirow{3}{*}{\textbf{EnDe}} & baseline        & 25.64 & - & 3464 & -\\
  & ASR \#25, chunk 1, context 300 & \textbf{38.46} & 66.59 & 3934 & 1522\\
& \cellcolor{lightgray}ASR gold, chunk 1, context 300 & \cellcolor{lightgray}39.84 & \cellcolor{lightgray}67.88 & \cellcolor{lightgray}2472 & \cellcolor{lightgray}0\\
 \hline
\multirow{3}{*}{\textbf{EnZh}} & baseline & 23.96 & - & 3275& -\\
 & ASR \#22, chunk 1, context 100 & \textbf{46.44} &  40.05 & 3698 & 1262\\
& \cellcolor{lightgray} 
 ASR \#11, chunk 3, context 100 & \cellcolor{lightgray}49.91 & \cellcolor{lightgray}43.08 & \cellcolor{lightgray}5449 & \cellcolor{lightgray}754 \\
 \hline
\multirow{2}{*}{\textbf{EnJa}}  &
 baseline & 16.19 & - & 3662 & - \\
& ASR \#22, chunk 2, context 200 & \textbf{34.69} & 42.89 & 4654 & 1262 \\
    
    \end{tabular}
    \caption{High-latency simultaneous translation results for English to German, Chinese, and Japanese on ACL6060 dev set compared to the IWSLT 2025 organizer's baseline and the contrastive systems (grey background).  The English-to-German contrastive system uses gold ASR. The English-to-Chinese one does not meet the SLAAL high latency limit of 4000 ms.}
    \label{tab:mt}
\end{table*}

\section{Conclusion}


In this paper, we presented our submission to the Simultaneous Speech Translation Task of the IWSLT 2025. Using the combination of the direct approach for Czech-to-English translation and the cascaded approach for  English to German, Chinese, and Japanese, we cover all language pairs of the task. To leverage the strong offline Whisper speech model and the large language model EuroLLM, we applied state-of-the-art onlinization techniques and further advancements such as prompting for context and domain adaptation. Our systems achieve a substantial improvement of 2 to 22 BLEU points over the IWSLT Organizers' baseline. Moreover, we propose a new robust approach to measure speech recognition latency.

\section*{Acknowledgements}
This paper has received funding from the
Project OP JAK Mezisektorová spolupráce
Nr. CZ.02.01.01/00/23\_020/0008518 named
``Jazykověda, umělá inteligence a jazykové a
řečové technologie: od výzkumu k aplikacím.''
The authors also acknowledge the support of
National Recovery Plan funded project MPO
60273/24/21300/21000 CEDMO 2.0 NPO.

\bibliography{how-real-biblio,msca-refs,iwslt25-bibs}

\appendix

\begin{table*}[ht]
    \centering
    \begin{tabular}{l|r|rrr}
 language  &  En & De & Zh & Ja \\
 \hline
 tokens per avg. recording (11.5 minutes) & 1\,963 & 2\,550 & 2\,423 & 2\,637  \\
 proportion of avg. recording in context & 1.04 & 0.91 & 0.93 & 0.87 \\
 max duration in context [minutes] & 11.96 & 10.5 & 10.7 & 10.0 \\
    \end{tabular}
    \caption{Estimation of maximum context duration of EuroLLM translation from English to German (De), Chinese (Zh), and Japanese (Ja), considering 11.5 minutes of average ACL6060 recording, and 4096 maximum context tokens containing the same content of the source and target.}
    \label{tab:dev-tokens}
\end{table*}

\section{Maximum Context Duration of EuroLLM}
\label{sec:maxcontext}

How long is the maximum context of EuroLLM in simultaneous mode, expressed in duration of long-form speech?

Consider the ACL6060 dev set reference in English, German, Japanese, and Chinese. It consists of five recordings with an average duration of 11.5 minutes. The average number of tokens per recording with the EuroLLM tokenizer for English, German, Japanese, and Chinese is in \Cref{tab:dev-tokens}. 

EuroLLM has a maximum context length of 4096 tokens. If the context contains parallel text in the source and target language, which is $x$-times English tokens plus $x$-times target tokens, and they sum up to 4096, $x$ is the maximum proportion of recording that fits into the context. Considering average recording, EuroLLM is able to fit a maximum of 10.5 minutes of English to German translation, 10.7 minutes of English to Chinese, or 10.0 minutes of English to Japanese. 





\end{document}